\documentclass[12pt,onecolumn,dvips]{article}
\usepackage{times}
\usepackage{epsfig}
\usepackage{graphicx}
\usepackage{amsmath}
\usepackage{amssymb}
\usepackage{algorithm}
\usepackage{algorithmic}

\DeclareMathOperator{\rank}{rank} \DeclareMathOperator{\tr}{tr}
\DeclareMathOperator{\diag}{diag} \DeclareMathOperator{\sgn}{sgn}
\renewcommand{\dfrac}{\displaystyle \frac}

\newtheorem{theorem}{Theorem}[section]

\newcommand{\pf}{\vspace{0.0em} \noindent {\bf Proof:}  }
\newcommand{\pfend}{\begin{flushright}
$\Box$
\end{flushright}
}

\begin{document}
\date{}
\title{\Large\bf Analysis and Improvement of Low Rank Representation for Subspace segmentation\footnote{Microsoft technical report \#MSR-TR-2010-177. This
work is done when S. Wei was visiting Microsoft Research Asia.}}
\author{Siming Wei$^1$ and Zhouchen Lin$^2$\\
{\normalsize$^1$Department of Computer science, Zhejiang
University, Hangzhou, P.R. China.} \\
{\normalsize(Email: tobiawsm@gmail.com)}\\
{\normalsize$^2$Microsoft Research Asia, Beijing, P.R. China.}\\
{\normalsize(Email: zhoulin@microsoft.com)}}

\maketitle \baselineskip 23pt

\begin{abstract}
We analyze and improve low rank representation (LRR), the
state-of-the-art algorithm for subspace segmentation of data. We
prove that for the noiseless case, the optimization model of LRR
has a unique solution, which is the shape interaction matrix (SIM)
of the data matrix. So in essence LRR is equivalent to
factorization methods. We also prove that the minimum value of the
optimization model of LRR is equal to the rank of the data matrix.
For the noisy case, we show that LRR can be approximated as a
factorization method that combines noise removal by column sparse
robust PCA. We further propose an improved version of LRR, called
Robust Shape Interaction (RSI), which uses the corrected data as
the dictionary instead of the noisy data. RSI is more robust than
LRR when the corruption in data is heavy. Experiments on both
synthetic and real data testify to the improved robustness of RSI.
\end{abstract}

\maketitle

\section{Introduction}
In many computer vision and machine learning problems, one often
assumes that the data is drawn from a union of multiple linear
subspaces. Thus subspace segmentation of such data has been
studied extensively. The existing methods for subspace
segmentation can be roughly divided into four groups: statistical
learning based methods (\cite{ma2007segmentation,
yang2006robust}), factorization based methods
(\cite{costeira1998multibody, gruber2004multibody,
vidal2008multiframe}), algebra based methods
\cite{ma2008estimation}, and sparsity based methods (e.g., SSC
\cite{elhamifar2009sparse} and LRR \cite{liu-robust}).

LRR \cite{liu-robust} is a recently proposed method and is
reported to have excellent performance on both synthetic and
benchmark data sets. For the noiseless case, LRR takes the data
itself as a dictionary and seeks the representation matrix with
the lowest rank. LRR can also handle noisy data by adding a
(2,1)-norm term to the objective function in order to make the
noise column sparse. The experimental results show that it is both
robust and accurate.

However, the motivation to utilize low rank criterion in LRR
remains vague. In \cite{liu-robust}, the authors only proved that
for noiseless case, the convex optimization model of LRR admits a
block diagonal solution, which is the representation they seek. As
the nuclear norm\footnote{The nuclear norm of a matrix is the sum
of the singular values of the matrix. It is the convex envelope of
the rank function on the unit matrix 2-norm ball and is often used
as an approximation of rank.} is not strongly convex, it is
unclear whether such a solution is unique and what it actually is.
If the uniqueness is not guaranteed, we may be at risk of finding
a non-block-diagonal solution. In this case the clustering
information will not be revealed. In the case of noisy data, the
authors simply added a (2,1)-norm term to the objective function
without providing sufficient insight to why such treatment can
work well.

\subsection*{Our Contributions}
In this paper, we present detailed analysis of LRR. We find that
although LRR was categorized by the authors of \cite{liu-robust}
as a sparsity based method, it is actually closely related to the
factorization methods. Our main contributions are:
\begin{enumerate}
\item We prove that in the noiseless case, LRR has a unique
solution, which is exactly the shape interaction matrix of the
data matrix. Consequently, the minimum objective function value is
the rank of the data matrix. \item For the noisy case, we show
that LRR can be roughly regarded as first applying a column sparse
robust PCA \cite{wright2009robust} to remove noise, and then
performing segmentation on the corrected data. \item We propose a
modified model, called Robust Shape Interaction (RSI) due to its
dependence on the shape interaction matrix, as an improvement of
LRR. Our experiments show that RSI is more robust and has better
performance than LRR.
\end{enumerate}

The remainder of this paper is organized as follows. Section
\ref{sec:Review} reviews the related state-of-the-art methods.
Section \ref{sec:Relationship} studies the relationship between
LRR and factorization methods. Then Section \ref{sec:RSI}
introduces RSI as an improvement of LRR. The simulation results on
synthetic and real data are shown in Section
\ref{sec:Experiments}. Finally, Section \ref{sec:Conclusions}
concludes the paper.

\section{Review of LRR and Factorization Methods}
\label{sec:Review}
\subsection{Basic Subspace Segmentation Problem}
Let $X=[x_1,x_2,...,x_n]$ be a collection of $m$ dimensional data
vectors drawn from a union of linear subspaces
$\{\mathcal{S}_i\}_{i=1}^k$, where the dimension of
$\mathcal{S}_i$ is $r_i$. The task of subspace segmentation (or
clustering) is to cluster the vectors in $X$ according to those
subspaces.

For notational simplicity, we may assume $X=[X_1,X_2,...,X_k]$,
where $X_i$ consists of the vectors in $\mathcal{S}_i$. For LRR
and factorization methods, it is assumed that the subspaces are
independent\footnote{If $\forall \alpha
\in\mbox{span}\{\mathcal{S}_1,\mathcal{S}_2,...,\mathcal{S}_k\}$,
the decomposition $\alpha=\sum_{i=1}^k{\alpha_i}$, where
$\alpha_i\in \mathcal{S}_i$, is unique, then we say that the
subspaces $\mathcal{S}_1,\mathcal{S}_2,...,\mathcal{S}_k$ are
independent.}.

%, i.e., $\sum_{i=1}^k\mathcal{S}_i=\oplus_{i=1}^k\mathcal{S}_i$.
Denote $d_i$ as the number of vectors in $X_i$. Then there must be
at least one block diagonal matrix
$Z=\diag\{Z_{1},Z_{2},...,Z_{k}\}$ satisfying
\begin{equation}
\label{XXZ}
X=XZ,
\end{equation}
where the size of the $i$-th block $Z_i$ is $d_i\times d_i$.
Equation (\ref{XXZ}) actually has an infinite amount of solutions.
Any solution is called a representation matrix. Note that the
block diagonal structure of $Z$ directly induces segmentation of
the data (each block corresponds to a cluster). So the clustering
task is equivalent to finding a block diagonal representation
matrix $Z$.

\subsection{Subspace Segmentation by Low-rank Representation}
As the solution to (\ref{XXZ}) is not unique, LRR
\cite{liu-robust} seeks the lowest rank representation matrix.
When the data is noiseless, LRR solves the following optimization
problem:
\begin{equation}
\label{LRR_noiseless}
\min_{Z} \|Z\|_{*}, \ \ \ \ s.t. \ \ X=XZ,
\end{equation}
where $\|\cdot\|_*$ denotes the nuclear norm. It was proved in
\cite{liu-robust} that the solution set of (\ref{LRR_noiseless})
includes at least a block diagonal representation matrix $Z^0$
that can be used for clustering.

%For the noisy data, the optimization model of LRR is formulated as:
%\begin{equation}
%\label{Convex_rep}
%\min_{Z,E} |Z|_*+\lambda|E|_{2,1} \ \ \ \ s.t., \ \  D=DZ, \ \ X=D+E,
%\end{equation}
%where $|E|_{2,1}=\sum_{j=1}^n\sqrt{\sum_{i=1}^n(E_{ij}^2)}$. Minimizing 2,1-norm of noise term is to meet the assumption that the corruptions are "sample specific"(\cite{liu-robust}), i.e, column sparse. Since (\ref{Convex_rep}) is non convex, LRR made an approximation that using $X$ itself as the dictionary instead of the clean data $D$. Then it
%targets to solve:
When the data is noisy, the optimization model of LRR is
formulated as:
\begin{equation}
\label{eqn:LRR_noisy} \min_{Z,E} \|Z\|_*+\lambda\|E\|_{2,1}, \ \ \
\ s.t. \ \  X=XZ+E,
\end{equation}
where $\|E\|_{2,1}=\sum_{j=1}^n\sqrt{\sum_{i=1}^n(E_{ij}^2)}$ is
the (2,1)-norm. Minimizing the (2,1)-norm of noise is to meet the
assumption that the corruptions are ``sample specific"
\cite{liu-robust}, i.e., some data vectors are corrupted and the
others are clean. Since in this case, the solution $Z^*$ to
(\ref{eqn:LRR_noisy}) may not be block diagonal, it is recognized
as an affinity matrix instead and spectral clustering methods are
applied to $|Z^*|+|(Z^*)'|$ to obtain a block diagonal matrix,
where $'$ denotes the matrix or vector transpose and $|A|$ denotes
a matrix whose entries are the absolute values of $A$.

\subsection{Factorization Methods and Shape Interaction Matrix}
Factorization based methods build a similarity matrix by
factorizing the data matrix and then applying spectral clustering
to the similarity matrix for clustering. This similarity matrix,
which is called the shape interaction matrix (SIM)
\cite{costeira1998multibody} in computer vision, is defined as
$SIM(X)=V_rV_r'$, where $X=U_rS_rV_r'$ is the skinny singular
value decomposition (SVD) of $X$ and $r$ is the rank of $X$. When
the data is noiseless, we have \cite{costeira1998multibody}:
\begin{theorem}
\label{thm:BLKDG} (Costeira and Kanade) Under the assumption that
the subspaces are independent and the data $X$ is clean, $SIM(X)$
is a block diagonal matrix that has exactly $k$ blocks. Moreover,
the $i$-th block on its diagonal is of size $d_i\times d_i$.
\end{theorem}
We can further have:
\begin{theorem}
\label{thm:rank} The rank of the $i$-th diagonal block of $SIM(X)$
is $r_i$.
\end{theorem}
\pf We partition $V_r'$ as
$V_r'=[V_{r,1}',V_{r,2}',...,V_{r,k}']$, where the number of
columns in $V_{r,i}'$ is $d_i$. Then $V_{r,i}V_{r,i}'$ is the
$i$-th diagonal block of $SIM(X)$ and $X_i=U_rS_rV_{r,i}'$. We can
also have
 $V_{r,i}'=S_r^{-1}U_r'X_i$. So $\rank(V_{r,i}')= \rank(X_i)=r_i$. The theorem is
 proved by using the relationship $\rank(V_{r,i}V_{r,i}')=\rank(V_{r,i}')$.
\pfend

Theorem \ref{thm:BLKDG} is the theoretical foundation of why SIM
can serve as the similarity matrix for subspace segmentation. When
the data contains noise, the SIM of the data matrix can still be
computed. Although in this case the SIM may not be block diagonal,
it can be made block diagonal, e.g., by applying spectral
clustering methods.

\section{Relationship between LRR and Factorization Methods}
\label{sec:Relationship} Though LRR was proposed as a sparsity
based method, we show in this section that it is equivalent to the
factorization methods. This is revealed by the following theorem:
\begin{theorem}
\label{thm:Equal_opt} The shape interaction matrix $SIM(X)$ is the
unique solution to the optimization problem (\ref{LRR_noiseless})
and the minimum objective function value is $\rank(X)$.
\end{theorem}
\pf
Let $[U_x,S_x,V_x]$ and $[U_z,S_z,V_z]$ be the full SVD of $X$
and $Z$, respectively. Denote $M=V_x'V_z$ and $N=V_x'U_z$. They
are both orthogonal matrices. Then $X=XZ$ is equivalent to
\begin{equation}\label{eqn:S_xM=S_xNS_z}
S_xM=S_xNS_z.
\end{equation}
Suppose $X$ is of rank $r$ and $M$ and $N$ are partitioned as
$M=\left(
\begin{array}{c}
M_r\\
M_{n-r}
\end{array}
\right)$ and $N=\left(
\begin{array}{c}
N_r\\
N_{n-r}
\end{array}
\right)$, respectively, where $M_r$ means that it consists of $r$
rows of $M$, etc. Then (\ref{eqn:S_xM=S_xNS_z}) reduces to
\begin{equation}\label{eqn:M_r=N_rS_z}
M_r=N_rS_z.
\end{equation}

Now consider (\ref{LRR_noiseless}). Since
\begin{equation}
\label{eqn:split1} NS_zM'=\left(
  \begin{array}{cc}
    N_rS_zM_r' & N_rS_zM_{n-r}'  \\
    N_{n-r}S_zM_r' & N_{n-r}S_zM_{n-r}'  \\
  \end{array}
  \right),
\end{equation}
by Lemma 3.1 in \cite{liu-robust}, we have
\begin{equation*}
\|N_rS_zM_r'\|_*\leq \|NS_zM'\|_*-\|N_{n-r}S_zM_{n-r}'\|_*.
\end{equation*}
As $\|N_rS_zM_r'\|_*=\|M_rM_r'\|_*=r$, where
(\ref{eqn:M_r=N_rS_z}) is applied, and
$\|NS_zM'\|_*=\|S_z\|_*=\|Z\|_*$, the above inequality reduces to
\begin{equation}
\label{eqn:opt_cond} r \leq \|Z\|_*-\|N_{n-r}S_zM_{n-r}'\|_* \leq
\|Z\|_*.
\end{equation}
Noticing $\|SIM(X)\|_*=r$ and $X=X\cdot SIM(X)$, we conclude that
the optimal $Z$ must satisfy $\|Z\|_*=r$, i.e., the minimum
objective function value is $\rank(X)$, and $SIM(X)$ is one of the
optimal solutions.

Next, we prove that $SIM(X)$ is the unique solution to
(\ref{LRR_noiseless}). Suppose that $Z_0$ is a solution to
(\ref{LRR_noiseless}). First, by (\ref{eqn:M_r=N_rS_z})
$N_{r}S_{z_0}M_{n-r}'=M_{r}M_{n-r}'=0$. Note that here and in the
sequel, $N_{r}$ and $M_{n-r}$ depend on $z_0$. Second, since
$\|Z_0\|_*=r$, from (\ref{eqn:opt_cond}) we have
$N_{n-r}S_{z_0}M_{n-r}'=0$. Thus (\ref{eqn:split1}) reduces to
\begin{equation}
\label{eqn:split2} NS_{z_0}M'=\left(
  \begin{array}{cc}
    N_rS_{z_0}M_r' & 0\\
    N_{n-r}S_{z_0}M_r' & 0 \\
  \end{array}
  \right),
\end{equation}
which implies the rank of $Z_0$ must be $r$, i.e., $S_{z_0}$ has
only $r$ nonzero entries on its diagonal. We then further
partition $M_r$ and $N_r$ as $M_r=[M_{r,r}, M_{r,n-r}]$ and
$N_r=[N_{r,r}, N_{r,n-r}]$, respectively, where $M_{r,r}$ consists
of $r$ columns of $M_r$, etc., and write
$S_{z_0}=\diag\{z_1,z_2,...,z_r,0,...,0\}$, where $z_1\geq z_2\geq
... \geq z_r>0$. Then by comparing both sides of
(\ref{eqn:M_r=N_rS_z}) we have $M_{r,n-r}=0$. So $M_{r,r}$ is an
orthogonal matrix and $M_{r,r}=N_{r,r}\diag\{z_1,z_2,...,z_r\}$.
We further denote $M_{r,r}^{r}$ and $N_{r,r}^{r}$ as the last
columns of $M_{r,r}$ and $N_{r,r}$, respectively. Then
$M_{r,r}^{r}=z_rN_{r,r}^{r}$ and thus we have
\begin{equation}
\label{greater1}
z_r=\frac{\|M_{r,r}^{r}\|_2}{\|N_{r,r}^{r}\|_2}=\frac{1}{\|N_{r,r}^{r}\|_2}\geq
1,
\end{equation}
where $\|\cdot\|_2$ denotes the 2-norm of a vector. The last
inequality holds because $N_{r,r}^{r}$ is part of a column of the
orthogonal matrix $N$. Since $z_r$ is the smallest nonzero
singular value of $Z_0$ and $\|Z_0\|_*=r$, (\ref{greater1})
implies that all the nonzero singular values of $Z_0$ must be 1.
So we conclude that $M_{r,r}=N_{r,r}$,
$S_{z_0}=\diag\{I_r,0_{n-r}\}$ and hence both $M$ and $N$ are
block diagonal. Finally,
\begin{eqnarray*}
\begin{array}{l}
Z_0=V_{z_0}S_{z_0}U_{z_0}'=(V_xN)S_{z_0}(V_xM)'=V_x(NS_{z_0}M')V_x'\\
=V_x\diag\{I_r,0_{n-r}\}V_x'=V_rV_r' =SIM(X).
\end{array}
\end{eqnarray*}
\pfend

As the uniqueness of the solution to LRR is guaranteed, we can
always use LRR to cluster clean data. Moreover, that the solution
is $SIM(X)$ can help us to understand the noisy LRR model
(\ref{eqn:LRR_noisy}).

\subsection*{Understanding the Noisy LRR Model}
Now consider the noisy case. It was not completely clear why
solving (\ref{eqn:LRR_noisy}) is effective in removing the column
sparse noise. Note that LRR uses the noisy data $X$ itself as the
dictionary instead of the clean data $D$, which is not quite
reasonable when the noise is heavy or the percentage of outliers
is relatively large. If we use the clean data as the dictionary,
the noisy LRR model (\ref{eqn:LRR_noisy}) would change to:
\begin{equation}
\label{eqn:Convex_rep} \min_{Z,D,E} \|Z\|_*+\lambda\|E\|_{2,1}, \
\ \ \ s.t. \ \  D=DZ, \ \ X=D+E.
\end{equation}
By Theorem \ref{thm:Equal_opt} it is straightforward to see that
(\ref{eqn:Convex_rep}) is equivalent to
\begin{equation}
\label{eqn:CRPCAZ} \min_{D,E}\, \rank(D)+\lambda\|E\|_{2,1}, \ \ \
s.t. \ \  X=D+E,
\end{equation}
and $Z=SIM(D)$. Model (\ref{eqn:CRPCAZ}) is very similar to the
robust PCA model \cite{wright2009robust}. It would decompose the
data into two parts: one is of low rank and the other is column
sparse. So we call problem (\ref{eqn:CRPCAZ}) column sparse robust
PCA (CSRPCA). Denote $\mathcal{S}=\oplus_{i=1}^k\mathcal{S}_i$.
Then the noise in the data can be decomposed into parts:
$E=E_s+E^{\perp}_s$, where each column of $E_s$ belongs to space
$\mathcal{S}$ and each column of $E^{\perp}_s$ is orthogonal to
$\mathcal{S}$. As rank minimization methods are to remove noise
outside the span of the clean data, CSRPCA is effective in
removing $E^{\perp}_s$ but is unable to remove $E_s$. However in
many real problems, the dimension of $\mathcal{S}$ (i.e., the rank
$r$ of clean data) is much smaller than the dimension of data. So
with high probability, $\|E_s\|_F$ is much smaller than
$\|E^{\perp}_s\|_F$, where $\|\cdot\|_F$ denotes the Frobenius
norm. Consequently, CSRPCA is able to remove most of the noise.

So we can see that the LRR model for noisy data is an
approximation of (\ref{eqn:Convex_rep}). Solving
(\ref{eqn:Convex_rep}) is equivalent to first applying CSRPCA to
remove the noise that is orthogonal to the space spanned by the
clean data, then computing the SIM to build the affinity matrix.
Since the noise level is greatly reduced, spectral clustering
methods on such an affinity matrix can be very effective to
segment the data into subspaces. This explains why LRR can have
good performance on noisy data.

\section{RSI: an Improved Version of LRR}
\label{sec:RSI} That $X$ itself can be used as the dictionary is
based on the assumption that the percentage of outliers is small
and the noise level is low. When this is not true, a more
reasonable formulation is (\ref{eqn:Convex_rep}), or equivalently
(\ref{eqn:CRPCAZ}). Following the convention, we may replace the
rank function in (\ref{eqn:CRPCAZ}) with the nuclear norm, giving
rise to the following convex optimization problem:
\begin{equation}
\label{CRPCAN}
\min_{D,E} \|D\|_{*}+\lambda\|E\|_{2,1}, \ \ \ s.t. \ \  X=D+E.
\end{equation}
Similar to robust PCA \cite{wright2009robust}, the above problem
can be solved by the inexact augmented Lagrange Multiplier
algorithm \cite{lin2009augmented}, which is based on the following
Lagrangian function:
\begin{eqnarray*}
\begin{array}{rcl}
L(D,E,Y)&\!\!=&\!\! \|D\|_* +\lambda\|E\|_{2,1}+ \left\langle {Y,X-D-E} \right\rangle \\
                 & & +\frac{\mu}{2}\|X-D-E\|_F^2,
\end{array}
\end{eqnarray*}
where $Y$ is the Lagrange multiplier, $\mu$ is a positive penalty
parameter, and $\langle A,B \rangle = \tr(A'B)$ is the inner
product of matrices.

When the ``clean" data $D$ is obtained, its representation matrix
can be obtained as $Z=SIM(D)$. As $D$ still contains noise $E_s$,
$Z$ may not be a block diagonal matrix. Like LRR and SSC
\cite{elhamifar2009sparse}, spectral clustering is also applied to
$|Z|$ to reveal the subspace clustering information. Note that as
$SIM(D)$ is symmetric, we do not have to use $|Z|+|Z'|$ as the
affinity matrix. We call this method Robust Shape Interaction
(RSI). The pseudo-code for RSI is presented in
Algorithm~\ref{alg:RSI}, where
$\Theta_\varepsilon(x)=\max(|x|-\varepsilon,0)\sgn(x)$ is the
thresholding operator. Readers are encouraged to refer to
\cite{lin2009augmented} and \cite{liu-robust} for the deduction of
the formulae in Algorithm~\ref{alg:RSI}.

Although SRI does not make a big change to LRR, as CSRPCA removes
most of the noise and SRI uses relatively clean data as the
dictionary, RSI is more robust than LRR, particularly when the
data is heavily corrupted.

\begin{algorithm}[tb]
   \caption{Robust Shape Interaction}
   \label{alg:RSI}
\begin{algorithmic}
   \STATE {\bfseries Input:} data matrix $X$, parameter $\lambda>0$.
   \STATE {\bfseries Initialize} : $E_0$, $Y_0$, $\mu_0>0$, $\mu_{\max}>\mu_0$, $\rho>1$, $\epsilon>0$.
   \STATE {\bfseries while} $\|X-D_k-E_k\|_\infty\geq \epsilon$ {\bfseries do}\\
   1. Update $D$ by solving $D_{k+1}=\arg\min\limits_{D}L(D,E_k,Y_k)=\arg\min\limits_{D} \|D\|_*+\frac{\mu_k}{2}\|X-D-E_k-\mu_k^{-1}Y_k\|_F^2$:
   $$(U,S,V)=\mbox{svd}(X-E_k-\mu^{-1}_kY_k),\ \ D_{k+1}=U\Theta_{\mu^{-1}_k}[S]V^{T}.$$\\
   2. Update $E$ by solving $E_{k+1}=\arg\min\limits_E L(D_{k+1},E,Y_k)=\arg\min\limits_E
   \lambda\|E\|_{2,1}+\frac{\mu_k}{2}\|X-D_{k+1}-E-\mu_k^{-1}Y_k\|_F^{2}$:\\
   Suppose the $i$-th column of $X-D_{k+1}-\mu_k^{-1}Y_k$ is $q_i$, then the $i$-th column of $E_{k+1}$ is $\Theta_{\lambda\mu_k^{-1}}(\|q_i\|_2)\dfrac{q_i}{\|q_i\|_2}$.\\
   3. Update $Y$ by: $Y_{k+1}=Y_{k}+\mu_k(X-D_{k+1}-E_{k+1}).$\\
   4. Update $\mu$ by: $\mu_{k+1}=\min(\rho\mu_k,\mu_{\max}).$ \\
   5. $k\leftarrow k+1.$  \\
  \STATE {\bfseries end while.}\\
  \STATE{\bfseries} Compute $Z=SIM(D)$.\\
  \STATE{\bfseries} Perform spectral clustering on $|Z|$.
  \STATE {\bfseries Output} : The subspace clusters indicated by the blocks of processed $|Z|$.
\end{algorithmic}
\end{algorithm}

\section{Experimental Results}
\label{sec:Experiments}
\subsection*{Clustering Synthetic Data}
\begin{figure}[h]
\center
    \includegraphics[width=4.0in]{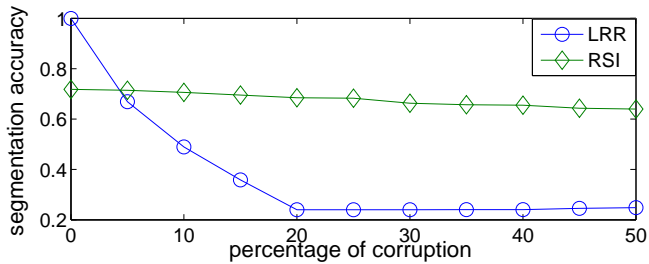}\vspace{-2mm}
    \caption{Segmentation accuracy of LRR and RSI. The parameters are set as
    $\lambda_{LRR}=0.12$ and
     $\lambda_{RSI}=0.6$, respectively.}
    \label{fig:LRR_vs_RSI}
\end{figure}

We first compare the robustness of LRR and RSI on synthetic data.
We construct 5 independent subspaces, each having a dimension of
4, and sample 20 data vectors with dimension 100 from each
subspace. For each data point $p$, small Gaussian noise of
variance $0.1*\|p\|_2$ is added. Moreover, we randomly choose a
certain percentage of points as outliers. For each outlier point
$p_{outlier}$ we add large Gaussian noise of variance
$\|p_{outlier}\|_2$ to it. Then we test LRR and RSI on this
corrupted data. The parameters are chosen as $\lambda_{LRR}=0.12$
and $\lambda_{RSI}=0.6$, respectively, which are both the optimal
for achieving the highest segmentation accuracy. For each
percentage of outliers, we repeat the experiment 20 times and
record the average accuracy and standard deviation. As shown in
Figure \ref{fig:LRR_vs_RSI}, RSI has better performance than LRR
when the percentage of outliers increases and its performance is
very stable. In this experiment, the maximum standard deviation of
RSI is 0.0403 and the average standard deviation is 0.0306. Thus
RSI is more robust than LRR on this synthetic data.

\subsection*{Clustering Real Data}
We then test the Hopkins155 motion database
\cite{tron2007benchmark} in this experiment. The database contains
156 video sequences and each of them is a clustering task. For
each sequence, there are $39\sim 550$ data vectors belonging to
two or three motions, each motion corresponding to a subspace. We
first repeat the same experiment, including the preprocessing, as
done in LRR (\cite{liu-robust}, Section 4.2) and then compare it
with RSI. The results are shown in Table I. Note that after
preprocessing, the data only contains slight corruptions. So both
RSI and LRR perform well. Evaluated by the average performance,
RSI outperforms LRR on this slightly corrupted dataset.
\begin{table}[h]
\label{tab:Hopkins}
\begin{center}
\begin{tabular}{|l|l|l|l|}
\hline
METHOD  \ \  & MEAN \ \  & MEDIAN \ \  & STD \ \ \\  %& MAX \ \  \\
\hline
LRR \ \  & 4.3673 \ \  & 0.4717 \ \  & 7.4540 \ \ \\ %& 33.0900 \ \  \\
RSI \ \  & 2.8501 \ \  & 0 \ \  & 7.5858 \ \ \\ %& 39.1549 \ \  \\
\hline
\end{tabular}
\end{center}
\caption{Segmentation error rate ($\%$) on the Hopkin155 database.
The parameters are selected as $\lambda_{LRR}=2.4$ and
$\lambda_{RSI}=0.24$, respectively.}
\end{table}

We finally test with the Extended Yale Database B
\cite{lee2005acquiring}, which consists of 640 frontal face images
of 10 subjects (there are 38 subjects in the whole database and we
use the first 10 subjects for our experiment). Each subject
contains about 64 images. The corruptions in this database is
heavy as more than half of the face images contain shadows or
specular lights. As did in LRR, we resize the images into
$48\times 42$ pixels and use the raw pixel values to form data
vectors of dimension 2016. The parameters are chosen as
$\lambda_{LRR}=0.05$ and $\lambda_{RSI}=0.6$ for LRR and RSI,
respectively, which are both optimal for achieving the lowest
segmentation error, based on our reimplemented code. The
clustering accuracies of LRR and RSI are 55.31\% and 58.13\%,
respectively. So RSI also performs better than LRR on this heavily
corrupted dataset.

\begin{figure}[t]
\center
    \includegraphics[width=3.7in]{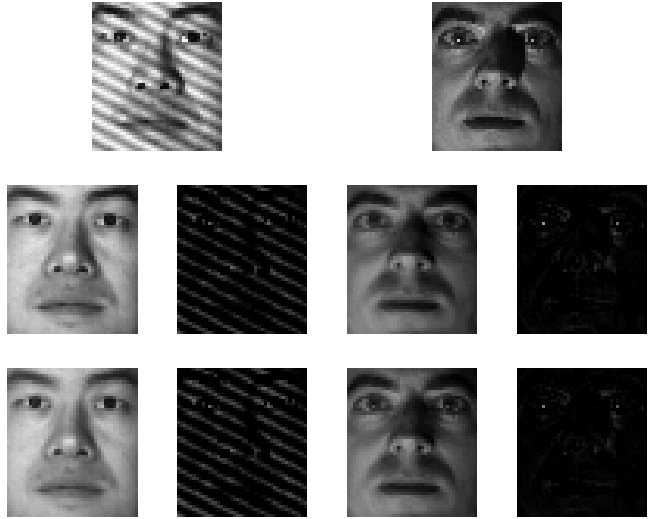}\vspace{-2mm}
    \caption{Two examples of using LRR and RSI to correct corruptions in face images in the Extended Yale Database B. The parameters are set as
    $\lambda_{LRR}=0.05$ and $\lambda_{RSI}=0.6$, respectively. The first row shows two original faces in the database.
     The second and third rows show the denoising result by LRR and RSI, respectively.
     For each face, the corrected images are displayed on the left, and the noise images are displayed on the right.}
    \label{fig:LRR_vs_RSI_Yale}
\end{figure}

Both LRR and RSI can be used to remove noise. Figure
\ref{fig:LRR_vs_RSI_Yale} shows two examples. From the figure, it
is clear that RSI is able to remove noise as well as LRR.

\section{Conclusions}\label{sec:Conclusions}
We have analyzed LRR and revealed its close relationship with the
factorization methods. In particular, we prove that when the data
is clean, the LRR problem has a unique solution, which is exactly
the shape interaction matrix of the data matrix, and the minimum
objective function value is the rank of the data matrix. We
further propose the Robust Shape Interaction method, which first
removes the noise by column sparse robust PCA, and then computes
the SIM of the corrected data for subspace segmentation. RSI is
verified by experiments to be more robust than LRR.

\bibliographystyle{IEEEtran}
\bibliography{RSI}

\begin{thebibliography}{10}
\providecommand{\url}[1]{#1}
\csname url@samestyle\endcsname
\providecommand{\newblock}{\relax}
\providecommand{\bibinfo}[2]{#2}
\providecommand{\BIBentrySTDinterwordspacing}{\spaceskip=0pt\relax}
\providecommand{\BIBentryALTinterwordstretchfactor}{4}
\providecommand{\BIBentryALTinterwordspacing}{\spaceskip=\fontdimen2\font plus
\BIBentryALTinterwordstretchfactor\fontdimen3\font minus
  \fontdimen4\font\relax}
\providecommand{\BIBforeignlanguage}[2]{{%
\expandafter\ifx\csname l@#1\endcsname\relax
\typeout{** WARNING: IEEEtran.bst: No hyphenation pattern has been}%
\typeout{** loaded for the language `#1'. Using the pattern for}%
\typeout{** the default language instead.}%
\else
\language=\csname l@#1\endcsname
\fi
#2}}
\providecommand{\BIBdecl}{\relax}
\BIBdecl

\bibitem{ma2007segmentation}
Y.~Ma, H.~Derksen, W.~Hong, and J.~Wright, ``Segmentation of multivariate mixed
  data via lossy data coding and compression,'' \emph{IEEE Transactions on
  Pattern Analysis and Machine Intelligence}, pp. 1546--1562, 2007.

\bibitem{yang2006robust}
A.~Yang, S.~Rao, and Y.~Ma, ``Robust statistical estimation and segmentation of
  multiple subspaces,'' in \emph{Computer Vision and Pattern Recognition
  Workshop on 25 Years on RANSAC}, 2006, pp. 99--107.

\bibitem{costeira1998multibody}
J.~Costeira and T.~Kanade, ``A multibody factorization method for independently
  moving objects,'' \emph{International Journal of Computer Vision}, vol.~29,
  no.~3, pp. 159--179, 1998.

\bibitem{gruber2004multibody}
A.~Gruber and Y.~Weiss, ``Multibody factorization with uncertainty and missing
  data using the {EM} algorithm,'' in \emph{IEEE Conference on Computer Vision
  and Pattern Recognition}, vol.~1, 2004, pp. 707--714.

\bibitem{vidal2008multiframe}
R.~Vidal, R.~Tron, and R.~Hartley, ``Multiframe motion segmentation with
  missing data using powerfactorization and {GPCA},'' \emph{International
  Journal of Computer Vision}, vol.~79, no.~1, pp. 85--105, 2008.

\bibitem{ma2008estimation}
Y.~Ma, A.~Yang, H.~Derksen, and R.~Fossum, ``Estimation of subspace
  arrangements with applications in modeling and segmenting mixed data,''
  \emph{SIAM Review}, vol.~50, no.~3, pp. 413--458, 2008.

\bibitem{elhamifar2009sparse}
E.~Elhamifar and R.~Vidal, ``Sparse subspace clustering,'' in \emph{IEEE
  Conference on Computer Vision and Pattern Recognition}, vol.~2, 2009, pp.
  2790--2797.

\bibitem{liu-robust}
G.~Liu, Z.~Lin, and Y.~Yu, ``Robust subspace segmentation by low-rank
  representation,'' in \emph{International Conference of Machine Learning},
  2010.

\bibitem{wright2009robust}
J.~Wright, A.~Ganesh, S.~Rao, and Y.~Ma, ``Robust principal component analysis:
  Exact recovery of corrupted low-rank matrices via convex optimization,''
  \emph{submitted to Journal of the ACM}, 2009.

\bibitem{lin2009augmented}
Z.~Lin, M.~Chen, L.~Wu, and Y.~Ma, ``The augmented {Lagrange} multiplier method
  for exact recovery of corrupted low-rank matrix,'' \emph{submitted to
  Mathematical Programming}, 2009.

\bibitem{tron2007benchmark}
R.~Tron and R.~Vidal, ``A benchmark for the comparison of {3-D} motion
  segmentation algorithms,'' in \emph{IEEE Conference on Computer Vision and
  Pattern Recognition}, 2007, pp. 1--8.

\bibitem{lee2005acquiring}
K.~Lee, J.~Ho, and D.~Kriegman, ``Acquiring linear subspaces for face
  recognition under variable lighting,'' \emph{IEEE Transactions on Pattern
  Analysis and Machine Intelligence}, pp. 684--698, 2005.

\end{thebibliography}

\end{document}